\documentclass[conference]{IEEEtran}
\IEEEoverridecommandlockouts
\usepackage{cite}
\usepackage{amsmath,amssymb,amsfonts}
\usepackage{graphicx}
\usepackage{textcomp}
\usepackage{xcolor}
\usepackage{url}
\usepackage{multirow}
\usepackage{arydshln}
\def\BibTeX{{\rm B\kern-.05em{\sc i\kern-.025em b}\kern-.08em
    T\kern-.1667em\lower.7ex\hbox{E}\kern-.125emX}}
\begin{document}

\title{Beyond Geometry: Efficient Topologically-Grounded Navigation in Complex 3D Environments}
\author{
\IEEEauthorblockN{1\textsuperscript{st} Yifan Du}
\IEEEauthorblockA{\textit{School of Integrated Circuits} \\
\textit{Sun Yat-sen University}\\
Shenzhen, China \\
duyf23@mail2.sysu.edu.cn}
\and
\IEEEauthorblockN{2\textsuperscript{nd} Chengwei Zhang}
\IEEEauthorblockA{\textit{School of Integrated Circuits} \\
\textit{Sun Yat-sen University}\\
Shenzhen, China \\
17858869215@163.com}
\and
\IEEEauthorblockN{3\textsuperscript{rd} Siyu Liao*}
\IEEEauthorblockA{\textit{School of Integrated Circuits} \\
\textit{Sun Yat-sen University}\\
Shenzhen, China \\
liaosy36@mail.sysu.edu.cn}
\and
\IEEEauthorblockN{4\textsuperscript{th} Zhongfeng Wang*}
\IEEEauthorblockA{\textit{School of Integrated Circuits} \\
\textit{Sun Yat-sen University}\\
Shenzhen, China \\
wangzf83@mail.sysu.edu.cn}
}

\maketitle

\begin{abstract}
Ground robot navigation in complex 3D environments is often hindered by geometric ambiguity, where non-traversable structures such as furniture share local geometric properties with navigable ground. Furthermore, the computational cost of searching massive voxel spaces remains a significant challenge. To address these issues, we present a surface extraction framework that constructs a reduced state space of physically reachable standing positions by enforcing ground support, overhead clearance, and seed-based connectivity constraints. Evaluation across five Matterport3D indoor scenes and three PCT benchmark scenes demonstrates over 80\% state space reduction and sub-millisecond A* search on the Matterport3D scenes, with 100\% planning success across all 300 tested queries.
\end{abstract}


\section{Introduction}

Deploying ground robots in unstructured 3D environments requires robust traversability identification. While modern reconstruction (LiDAR SLAM, RGB-D fusion) provides detailed maps, it introduces \textit{geometric ambiguity}: non-traversable structures such as furniture, ceiling cavities, and artifacts share local geometric properties (planarity, low slope) with navigable ground, making them indistinguishable to geometry-only planners.

Existing methods primarily rely on local geometric features: 3D2M~\cite{wang2023towards} applies slope and step-height filtering on point cloud neighborhoods; PCT~\cite{yin2024tomography} horizontally slices clouds into traversable layers. Neither can distinguish a tabletop from a floor or a ceiling cavity from a corridor.

We resolve this ambiguity with a single additional constraint: \textit{physical reachability from the robot's current position}. We extract the connected component of geometrically valid candidates reachable from a known seed via bounded-height steps---topologically filtering all disconnected surfaces (furniture, artifacts, inter-floor cavities) regardless of geometry. Restricting the state space to this reachable surface ($<$20\% of the full grid in furniture-rich indoor scenes) yields substantial computational savings.

Building on this observation, we develop a complete framework from surface extraction to clearance-aware planning, with the following contributions:
\begin{enumerate}
    \item A surface extraction algorithm that constructs a seed-reachable traversable state space via topological filtering, eliminating geometrically ambiguous regions without semantic labels, under the static-map assumption.
    \item Systematic evaluation across eight scenes from two public benchmarks (Matterport3D and PCT~\cite{yin2024tomography}), demonstrating that seed-based connectivity alone achieves over 80\% state space reduction in furniture-rich indoor environments and enables sub-millisecond planning without per-scene parameter tuning.
\end{enumerate}

Section~\ref{sec:extraction_results} validates state-space reduction after extraction, Section~\ref{sec:planning_eval} evaluates planning success and search latency on Matterport3D, and Section~\ref{sec:pct_comparison} reports cross-scenario comparisons on PCT benchmarks.

\section{Related Work}

\subsection{Geometric Ambiguity in Traversability Estimation}
Geometry-only planners classify traversability using local features such as slopes, step-heights, and surface normals. The 3D2M-planner~\cite{wang2023towards} identifies ground layers via local slope and step-height filtering (RANSAC-based Valid Ground Filter) applied to point cloud neighborhoods; ugv\_nav4d~\cite{b5} evaluates terrain traversability on multi-level surface (MLS) maps~\cite{b4}. These methods produce false positives on furniture surfaces and floating artifacts that locally mimic floor geometry~\cite{b7,b8}. Point cloud tomography~\cite{yin2024tomography} horizontally slices point clouds into multi-level 2.5D traversable layers, achieving highly efficient planning in large-scale outdoor structures with well-separated vertical layers. When applied to cluttered indoor scenes, however, horizontal slicing can include furniture and artifacts at valid floor elevations, since it does not enforce physical reachability from a known position.

Semantic traversability methods (e.g., learning-based terrain classifiers) can in principle distinguish floors from furniture, but require dense labeled training data and high-fidelity perception pipelines, limiting deployment in previously unseen or resource-constrained settings. Our seed-based connectivity addresses the geometric ambiguity failure mode through a purely geometric-topological criterion---requiring that every traversable state be reachable from the robot's position via bounded-height steps---with minimal computational overhead and no scene-specific training.

\subsection{Computational Efficiency in 3D Navigation}
Conventional volumetric approaches~\cite{hornung2013octomap, putz2016navmesh, wang2023towards} process the complete 3D grid at a cost fixed by the grid size $V = N_x N_y N_z$. Because these methods lack a connectivity-aware filtering mechanism, they exhaustively evaluate the entire space, irrespective of how much of the volume is actually reachable. By restricting planning to the extracted surface $\mathcal{S}$ ($|\mathcal{S}| \ll V$), our approach reduces search cost as a direct consequence of the state space reduction.

\section{Methodology}

\subsection{Problem Formulation}
\label{sec:formulation}
Consider an environment represented as a 3D occupancy grid $\mathcal{O}: \mathbb{Z}^3 \to \{0,1\}$ at resolution $r$, typically reconstructed from LiDAR SLAM, RGB-D fusion, or offline methods such as 3DGS~\cite{b9}. The grid contains both traversable structures (floors, stairs) and non-traversable ones (furniture, ceilings) that share identical local geometric properties, making geometry alone insufficient for traversability identification.

The objective is to extract $\mathcal{S} \subset \mathbb{Z}^3$---the set of free voxels where a robot can physically stand---such that every element of $\mathcal{S}$ is reachable from a known seed position, ensuring traversability without post-hoc validation. In practice, the seed is obtained by discretizing the robot pose onto the occupancy grid and selecting the nearest valid candidate voxel in $\mathcal{C}$.

Prior to surface extraction, obstacle inflation expands occupied voxels by the robot's collision radius $R_{\text{inf}}$ in the XY plane ($\pm 1$ voxel vertically), ensuring collision safety without blocking valid passages under low overhangs.

\subsection{Surface Extraction}
\label{sec:extraction}
We propose a surface extraction algorithm that reliably identifies traversable positions in cluttered indoor environments. Our approach applies geometric preconditions followed by a topological reachability constraint that resolves cases where geometry alone is insufficient.

\subsubsection{Geometric Constraints}
Using the occupancy grid $\mathcal{O}$ defined in Section~\ref{sec:formulation}, a voxel $(x, y, z)$ is a \textit{surface candidate} if it satisfies two geometric conditions. Let $K = \lceil H_{\text{clear}}/r \rceil$ denote the clearance height in voxels.

\textbf{Condition 1 (Ground Support):} The voxel must be free and rest on a solid surface directly below:
\begin{equation}
    \mathcal{O}(x, y, z) = 0 \quad \land \quad \mathcal{O}(x, y, z-1) = 1
\end{equation}
This filters wall surfaces, suspended objects (e.g., hanging lamps), and mid-air reconstruction artifacts.

\textbf{Condition 2 (Overhead Clearance):} A vertical column of $K$ voxels above must be free to accommodate the robot's height:
\begin{equation}
    \forall j \in \{1, \dots, K\}: \mathcal{O}(x, y, z+j) = 0
\end{equation}
This filters non-traversable spaces under low furniture. Let $\mathcal{C}$ denote the candidate set of all voxels satisfying both conditions.

\subsubsection{Topological Constraint: Seed-Based Connectivity}
Geometric constraints alone cannot filter furniture surfaces---a tabletop satisfies both ground support and overhead clearance and thus appears in $\mathcal{C}$. The critical additional constraint is \textit{physical reachability}.

Let $k = \lfloor T_{\text{conn}}/r \rfloor$ denote the maximum step height in voxels, where $T_{\text{conn}}$ is the traversable step threshold (e.g., $0.3$\,m). Two candidates $\mathbf{c}_1, \mathbf{c}_2 \in \mathcal{C}$ are \textit{Z-connected neighbors} ($\mathbf{c}_1 \sim_Z \mathbf{c}_2$) if
\begin{equation}
    \|\Delta \mathbf{c}_{xy}\|_1 = 1 \;\;\land\;\; |\Delta z| \le k
\end{equation}
where $\Delta \mathbf{c}_{xy}$ and $\Delta z$ denote horizontal and vertical displacements, respectively. The traversable surface $\mathcal{S}$ is the connected component of $\mathcal{C}$ containing the seed $\mathbf{s}_0$ under $\sim_Z$, constructed via BFS.

This connectivity constraint isolates furniture (floor-to-tabletop discontinuity $> k$), ceiling cavities (no reachable path from floor seeds), and isolated reconstruction artifacts. The seed $\mathbf{s}_0$ is naturally the robot's current pose; multi-seed BFS supports incremental exploration.

\subsubsection{Multi-Level Support}
The BFS naturally handles multi-story structures: each stair step satisfies $|\Delta z| \le k$, enabling propagation across floors.

A key property is the multi-valued vertical mapping: a single horizontal coordinate $(x,y)$ may correspond to multiple valid heights,
\begin{equation}
    \mathcal{S}_{x,y} = \{ z \mid (x, y, z) \in \mathcal{S} \}
\end{equation}
where $|\mathcal{S}_{x,y}| > 1$ in multi-story buildings and overpass structures---configurations that 2.5D elevation maps~\cite{b2} cannot represent. This property requires the path planner to operate on full 3D states $(x,y,z) \in \mathcal{S}$ rather than 2D projections, ensuring floors at different heights are never conflated.

\begin{figure}[t]
\centering
\includegraphics[width=\columnwidth]{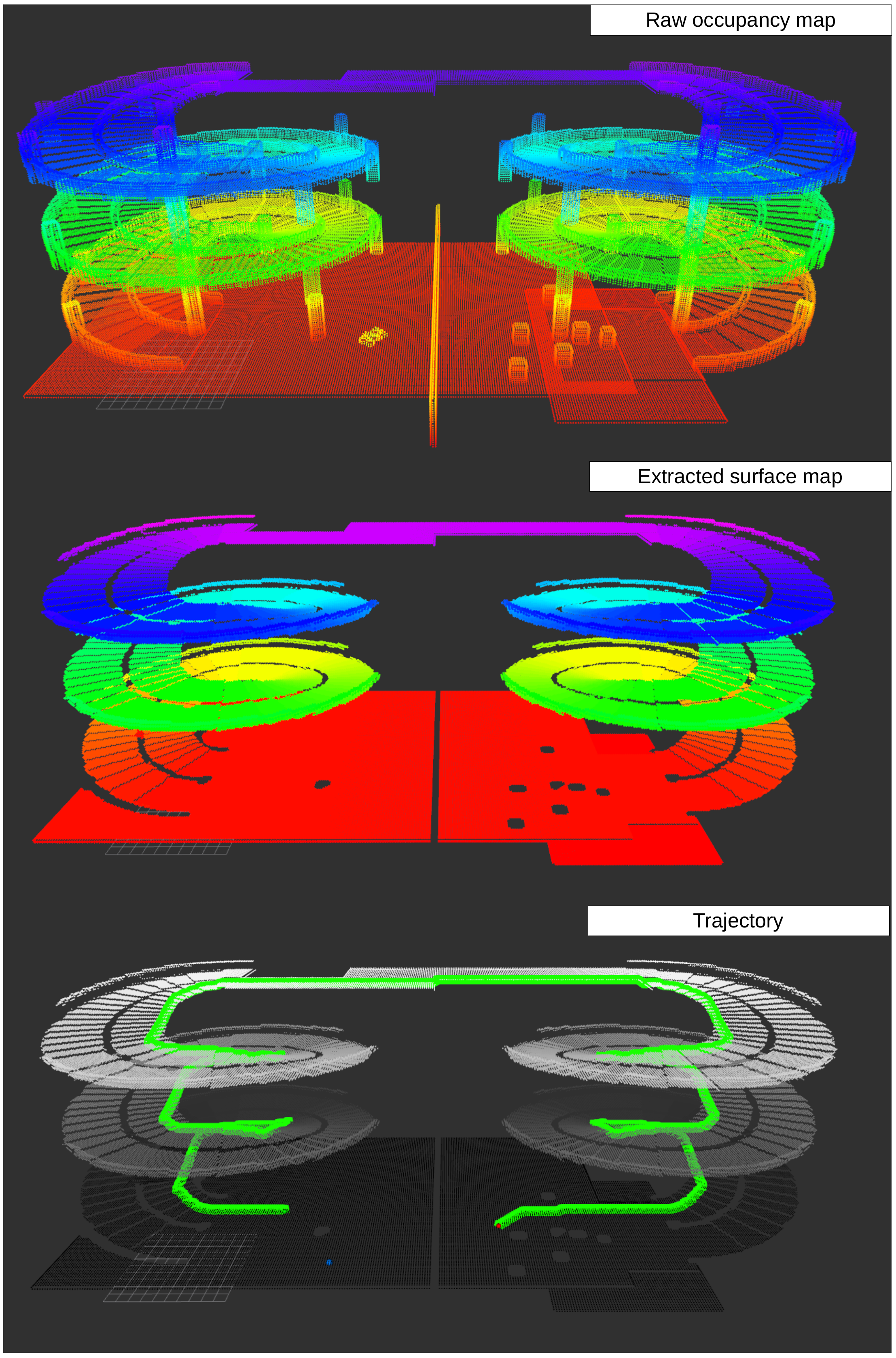}
\caption{Pipeline overview. (a) Raw 3D occupancy map. (b) Extracted surface $\mathcal{S}$: seed-based filtering retains only reachable floors and stairs. (c) Cross-floor trajectory planned on $\mathcal{S}$.}
\label{fig:surface_extraction}
\end{figure}

Table~\ref{tab:filtering} and Figure~\ref{fig:surface_extraction} collectively demonstrate the filtering mechanisms across various indoor scenarios, highlighting the removal of geometrically ambiguous structures.

\begin{table}[b]
\centering
\caption{Filtering effects of the three surface constraints. ``--'': constraint is not reached (already filtered by a prior constraint).}
\label{tab:filtering}
\begin{tabular}{lccc}
\hline
\textbf{Structure} & \textbf{Ground} & \textbf{Clearance} & \textbf{Connectivity} \\ \hline
Floor (valid) & \checkmark & \checkmark & \checkmark \\
Stairs (valid) & \checkmark & \checkmark & \checkmark \\
Wall surface & $\times$ & -- & -- \\
Ceiling interior & \checkmark & \checkmark & $\times$ \\
Tabletop & \checkmark & \checkmark & $\times$ \\
Under-table space & \checkmark & $\times$ & -- \\
Reconstruction hole & \checkmark & \checkmark & $\times$ \\ \hline
\end{tabular}
\end{table}

\subsection{Path Planning}
With $\mathcal{S}$ extracted, planning reduces to graph search on $\mathcal{S}$. The speedup over volumetric search is a direct consequence of state space reduction ($|\mathcal{S}| \ll V$), not algorithmic novelty.

We employ weighted A* with $f(\mathbf{s}) = g(\mathbf{s}) + \epsilon \cdot h(\mathbf{s})$ ($\epsilon \ge 1$). Successors of $\mathbf{s}$ are its Z-connected neighbors in the four cardinal (4-connected) directions, uniquely identified by full 3D coordinates $(x, y, z) \in \mathcal{S}$ to prevent floor aliasing in multi-story environments. Let $D(\mathbf{s})$ denote the 4-connected Manhattan distance from $\mathbf{s}$ to the nearest occupied boundary within $\mathcal{S}$, precomputed by a two-pass scan~\cite{rosenfeld1966} at $O(|\mathcal{S}|)$ cost. The edge cost is:
\begin{equation}
    c(\mathbf{s}, \mathbf{s}') = r\|\mathbf{s}' \!-\! \mathbf{s}\|_2 + |\Delta z| \cdot r \cdot \omega_z + \frac{\omega_{\text{obs}} \cdot r}{D(\mathbf{s}') + 1}
\end{equation}
where $\omega_z \in \{\omega_\uparrow, \omega_\downarrow, 0\}$ penalizes height changes asymmetrically (ascent $>$ descent), and the last term biases paths away from obstacles. The admissible heuristic is:
\begin{equation}
    h(\mathbf{s}) = r\|\mathbf{s} - \mathbf{s}_{\text{goal}}\|_2 + r|\Delta z_g| \omega_\downarrow
\end{equation}
which is admissible since it underestimates all cost terms.

The planner outputs a discrete path directly from weighted A* on $\mathcal{S}$; no post-search trajectory optimization is applied in the experiments reported here. Accordingly, the Matterport3D evaluation reports mean A* search time $T$, while the PCT-style single-query comparison separately reports map build time $T_p$, candidate evaluation time $T_e$, search time $T_s$, and total pipeline time $T_{\text{all}} = T_p + T_e + T_s$.

\section{Experiments}

\subsection{Experimental Setup}
\label{sec:setup}
We evaluate on two complementary benchmark suites:

\textbf{Matterport3D scenes} (S1--S5). Five indoor scenes from the Matterport3D dataset~\cite{b6}. These span multi-story floors, dense furniture, reconstruction artifacts, and multi-valued vertical geometry. Surface extraction experiments and the Matterport3D planning evaluation use this suite exclusively.

\textbf{PCT benchmark scenes} (P1--P3). Three publicly available 3D point cloud scenes released with the PCT planner~\cite{yin2024tomography} and 3D2M-planner~\cite{wang2023towards}: (P1)~Building---a multi-floor indoor structure with stairs and slopes, $(22\!\times\!20\!\times\!16)$\,m; (P2)~Plaza---an outdoor urban scenario with various structures, $(56\!\times\!56\!\times\!5)$\,m; (P3)~Spiral---a large-scale structure with continuous spiral ramp, $(80\!\times\!40\!\times\!23)$\,m. P1 and P2 enable direct comparison against PCT~\cite{yin2024tomography} and 3D2M~\cite{wang2023towards} on shared data and query points; P3 is used for state space analysis only.

\textbf{Implementation.} All experiments were run on an AMD Ryzen 7 9800X3D CPU. Extraction parameters: $T_{\text{conn}}\!=\!0.3$\,m, $H_{\text{clear}}\!=\!1.6$\,m, $R_{\text{inf}}\!=\!0.3$\,m. For Matterport3D scenes (S1--S5), $r\!=\!0.2$\,m. For PCT benchmark scenes, we adopt the resolutions reported in~\cite{yin2024tomography} to ensure a fair comparison: $r\!=\!0.1$\,m for P1~(Building), P2~(Plaza), and P3~(Spiral). Planning parameters: $\epsilon\!=\!1$ (optimal A*), $\omega_\uparrow\!=\!2.0$, $\omega_\downarrow\!=\!1.0$, $\omega_{\text{obs}}\!=\!0.5$.

\textbf{Time normalization.} PCT~\cite{yin2024tomography} reports timing on an Intel i9-12900KF. We define the hardware ratio $\alpha \!=\! 3338/2586 \!\approx\! 1.29$ from Geekbench~6 single-core scores~\cite{geekbench_9800x3d,geekbench_i9}; multiplying our CPU timings by $\alpha$ gives an approximate equivalent i9-12900KF performance. This normalization is used as an indicative CPU-side correction only, while speed-independent metrics (path length, success rate) are compared directly.

\textbf{Baseline implementations.} For the Matterport3D evaluation, we run the official 3D2M codebase~\cite{wang2023towards} as the baseline. For the PCT benchmark comparisons, PCT and 3D2M results are taken from the published values reported in~\cite{yin2024tomography}.

\subsection{Surface Extraction and State Space Reduction}
\label{sec:extraction_results}

Table~\ref{tab:extraction} reports surface extraction results across all eight scenes.

\begin{table}[b]
\centering
\caption{Surface extraction results. $V$: total voxels (K); $|\mathcal{S}|$ (K): extracted traversable states; Reduction: $1 - |\mathcal{S}|/V$; $T_{\text{ext}}$: extraction time.}
\label{tab:extraction}
\resizebox{\columnwidth}{!}{%
\begin{tabular}{lcr r r r}
\hline
\textbf{Scene} & \textbf{Dims (m)} & $V$ \textbf{(K)} & $|\mathcal{S}|$ \textbf{(K)} & \textbf{Reduction (\%)} & \textbf{$T_{\text{ext}}$ (ms)} \\
\hline
S1: 1LXtFkjw3qL & $12.2\!\times\!22.2\!\times\!12.2$ & 63003 & 8587 & 86.3 & 1.15 \\
S2: 1pXnuDYAj8r      & $20\!\times\!25.2\!\times\!7$   & 59690 & 8073 & 86.4 & 1.07 \\
S3: 2azQ1b91cZZ      & $43.2\!\times\!25.8\!\times\!8.6$   & 82949 & 8657 & 89.5 &  1.36 \\
S4: 2n8kARJN3HM   & $24.6\!\times\!26\!\times\!9$   & 92333 & 12046 & 86.9 & 1.58 \\
S5: 5LpN3gDmAk7   & $25.2\!\times\!24.4\!\times\!8.8$   & 58930 & 10988 & 81.3 & 1.43 \\
\hdashline
P1: Building    & $22\!\times\!20\!\times\!16$  & 250842 & 81544 & 67.4 & 12.83 \\
P2: Plaza       & $56\!\times\!56\!\times\!5$   & 502410 & 211146 & 57.9 & 33.54 \\
P3: Spiral      & $80\!\times\!40\!\times\!23$  & 212609 & 119381 & 43.8 & 13.75 \\
\hline
\end{tabular}}
\end{table}

Across S1--S5, state space reduction exceeds 80\%; extraction completes under 2\,ms for all Matterport3D scenes---well within online replanning budgets---and under 35\,ms on the larger PCT scenes, still within typical planning cycle budgets.

\begin{figure}[t]
\centering
\includegraphics[width=\columnwidth]{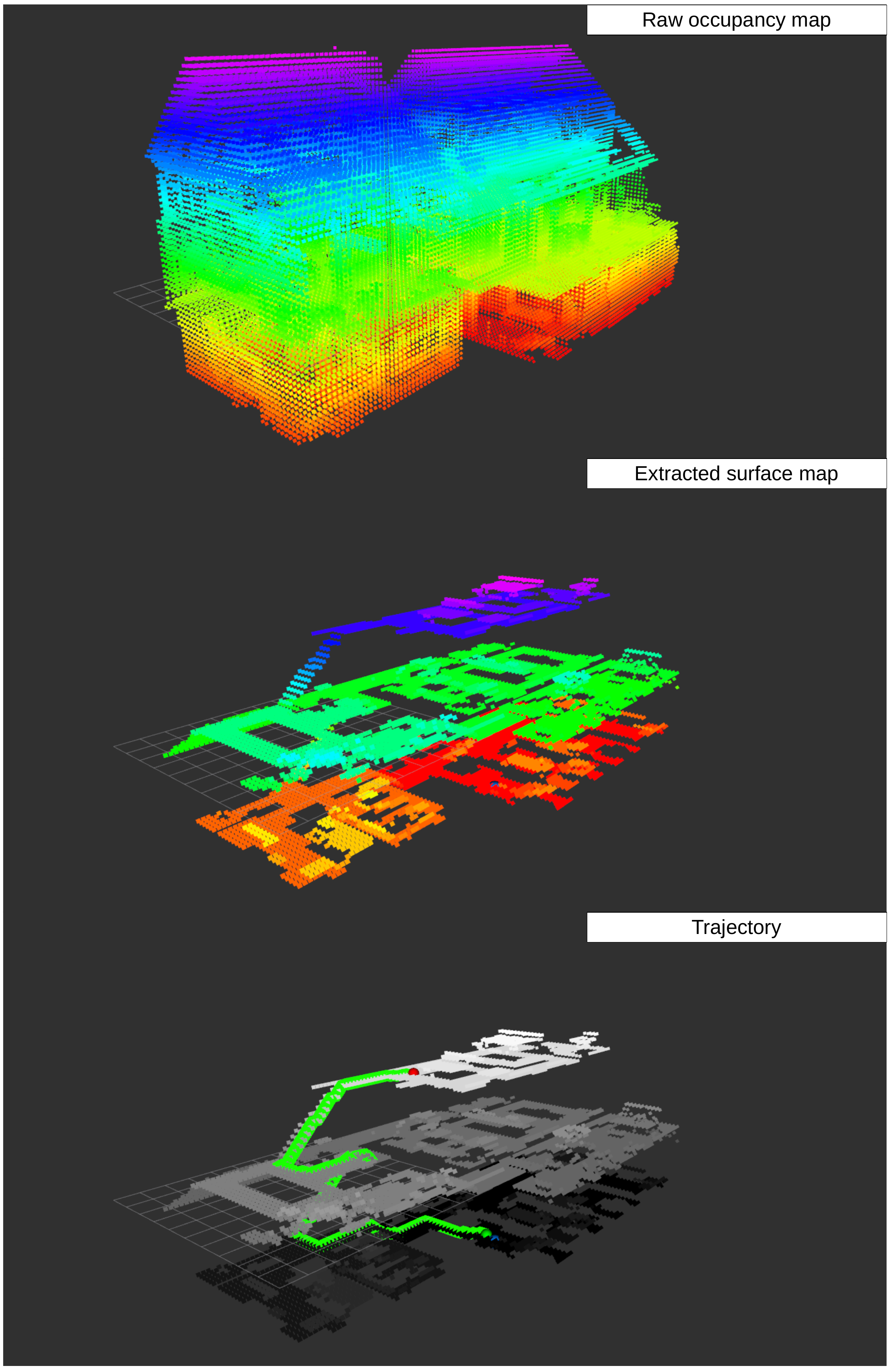}
\caption{Scene S1: (a) raw occupancy map; (b) extracted surface $\mathcal{S}$ (Table~\ref{tab:extraction})---only reachable floors and stairs retained; (c) cross-floor planned trajectory.}
\label{fig:extraction_result}
\end{figure}

Figure~\ref{fig:extraction_result} shows Scene S1: after extraction, $\mathcal{S}$ retains only reachable floors and stairs (86.3\% reduction), and a cross-floor query is solved in sub-millisecond time.

\subsection{Path Planning Evaluation on Matterport3D}
\label{sec:planning_eval}

Table~\ref{tab:planning} reports per-scene planning results across S1--S5 using 50 start-goal queries per scene (25 same-floor, 25 cross-floor), totaling 250 queries. Start and goal positions are uniformly sampled from $\mathcal{S}$; same-floor queries share the same height layer, while cross-floor queries span different floors. We compare against the official 3D2M implementation~\cite{wang2023towards} (see Section~\ref{sec:setup} for details).

\begin{table}[t]
\centering
\caption{Per-scene planning (50 queries each). SR: success rate; $T$: mean A* search time for our planner and the official 3D2M implementation (Section~\ref{sec:setup}); $L$: mean discrete path length.}
\label{tab:planning}
\resizebox{\columnwidth}{!}{%
\begin{tabular}{l cc cc cc}
\hline
 & \multicolumn{2}{c}{\textbf{SR (\%)}} & \multicolumn{2}{c}{$T$ \textbf{(ms)}} & \multicolumn{2}{c}{$L$ \textbf{(m)}} \\
\cline{2-3} \cline{4-5} \cline{6-7}
\textbf{Scene} & \textbf{Ours} & \textbf{3D2M} & \textbf{Ours} & \textbf{3D2M} & \textbf{Ours} & \textbf{3D2M} \\
\hline
S1: 1LXtFkjw3qL & 100 & 50 & 0.38 & 127.66 & 32.05 & 37.78 \\
S2: 1pXnuDYAj8r      & 100 & 67 & 0.19 & 179.42 & 31.37 & 32.16 \\
S3: 2azQ1b91cZZ      & 100 & 87 & 0.25 & 164.45 & 35.17 & 31.81 \\
S4: 2n8kARJN3HM   & 100 & 60 & 0.51 & 284.80 & 41.26 & 20.10 \\
S5: 5LpN3gDmAk7   & 100 & 67 & 0.65 & 148.55 & 48.79 & 28.85 \\
\hdashline
\textbf{Average}& 100 & 66 & 0.40 & 180.97 & 37.72 & 30.14 \\
\hline
\end{tabular}}
\end{table}

Our planner achieves 100\% success versus 66\% for the official 3D2M baseline, whose failures in our evaluation all involved cross-floor queries where geometry-only filtering cannot identify a valid inter-floor path. Mean A* search time ranges from 0.19\,ms to 0.65\,ms (average 0.40\,ms), a $452{\times}$ speedup compared to 3D2M's 181.0\,ms; even including extraction overhead ($\sim$1.3\,ms, Table~\ref{tab:extraction}), the total pipeline remains over $100{\times}$ faster. These speedups are a direct consequence of searching the reduced surface $\mathcal{S}$ rather than the full voxel grid.

\subsection{Cross-Scenario Comparison on PCT Benchmarks}
\label{sec:pct_comparison}

We evaluate on P1--P2 against PCT~\cite{yin2024tomography} and 3D2M~\cite{wang2023towards}, using the published CPU-only timings from~\cite{yin2024tomography} (only GPU code is released, precluding independent reproduction), adjusted by $\alpha$ (Section~\ref{sec:setup}).

\subsubsection{Single-Query Comparison}
Table~\ref{tab:pct_single} reports results on P1 (Building) using the fixed start--goal pair $(5,5)\!\to\!(-6,-1)$ from~\cite{yin2024tomography}.

\begin{table}[t]
\centering
\caption{Single-query efficiency on P1~(Building), following~\cite{yin2024tomography} Table~III. $T_p$: map build; $T_e$: evaluation; $T_s$: search; $N_s$: nodes; $T_o$: optimization; $T_{\text{all}}$: total. Baselines are copied from~\cite{yin2024tomography} (i9-12900KF); ours is measured on Ryzen~7 9800X3D ($\alpha\!\approx\!1.29$). ``--'': not applicable.}
\label{tab:pct_single}
\resizebox{\columnwidth}{!}{%
\begin{tabular}{llrrrrrrr}
\hline
\textbf{Scene} & \textbf{Method} & $T_p$ & \textbf{Size} & $T_e$ & $T_s$ & $N_s$ & $T_o$ & $T_{\text{all}}$ \\
 & & \textbf{(ms)} & \textbf{(MB)} & \textbf{(ms)} & \textbf{(ms)} & & \textbf{(ms)} & \textbf{(ms)} \\
\hline
\multirow{5}{*}{P1: Building}
  & Ours                           & \textbf{38.02} & 17.10 & \textbf{18.21}           & 7.90 & $73{,}469$        & --            & \textbf{64.13} \\
  & PCT~\cite{yin2024tomography}    & 290.98       & \textbf{3.17}         & 234.95        & \textbf{7.33}         & $39{,}760$          & 359.39        & 892.65 \\
  & 3D2M~\cite{wang2023towards}     & $1{,}910$    & 28.16        & $11{,}530$    & --           & --                  & --            & -- \\
  & P\"{u}tz~\cite{putz2021navmesh} & $3{,}410$    & 17.10        & $11{,}360$    & --           & --          & --            & --  \\
  & Liu~\cite{liu2016robotic}       & --           & 7.40         & $116{,}470$   & 765.54       & $2.09\!\times\!10^5$ & --           & $117{,}510$ \\
\hline
\end{tabular}}
\end{table}

Even after hardware normalization (multiplying our timings by $\alpha \approx 1.29$ to estimate equivalent i9-12900KF performance), our total pipeline time of $64.13 \times 1.29 \approx 83$\,ms on P1 remains more than $10{\times}$ faster than PCT's 893\,ms. The improvement stems primarily from faster map construction and candidate evaluation ($T_p\!+\!T_e$: 56\,ms vs.\ 526\,ms for PCT), while A* search times are comparable (7.90 vs.\ 7.33\,ms); our pipeline also eliminates the trajectory optimization step ($T_o\!=\!359$\,ms) that PCT requires.

\subsubsection{Multi-Query Evaluation on Plaza}
To assess consistency across diverse goal locations, we follow~\cite{yin2024tomography} and uniformly sample 50 goals at a 26.5\,m radius from the map center $(0,0)$ on P2 (Plaza).
Table~\ref{tab:pct_plaza50} summarizes the search-time, path-length, and success-rate statistics for this 50-query setting.

\begin{table}[t]
\centering
\caption{Multi-query evaluation on Plaza (P2), 50 goals at 26.5\,m radius. $T_s$: search time; $L_t$: trajectory length; SR: success rate. All baseline data as published in~\cite{yin2024tomography} (i9-12900KF); our timings on AMD Ryzen 7 9800X3D ($\alpha \approx 1.29$, see Section~\ref{sec:setup}).}
\label{tab:pct_plaza50}
\begin{tabular}{lccc}
\hline
\textbf{Method} & $T_s$ \textbf{(ms)} & $L_t$ \textbf{(m)} & \textbf{SR (\%)} \\
\hline
Ours                         & $2.74\!\pm\!1.06$            & $30.11\!\pm\!2.31$           & 100 \\
PCT~\cite{yin2024tomography} & $13.93\!\pm\!3.93$      & $28.32\!\pm\!1.55$        & 100 \\
3D2M~\cite{wang2023towards}  & $1{,}190\!\pm\!335$     & $47.13\!\pm\!9.59$       & 94 \\
P\"{u}tz~\cite{putz2021navmesh} & $44.16\!\pm\!4.43$   & $30.69\!\pm\!1.53$          & 92 \\
Liu~\cite{liu2016robotic}    & $358.2\!\pm\!25.8$      & $28.91\!\pm\!0.71$          & 12 \\
\hline
\end{tabular}
\end{table}

As shown in Table~\ref{tab:pct_plaza50}, our method attains 100\% success on the Plaza benchmark with the lowest reported search time among the compared methods, while maintaining path lengths close to PCT and P\"{u}tz. The comparable path quality confirms that connectivity-based filtering does not sacrifice route efficiency, while the timing advantage is consistent across both single-query and multi-query settings.

\section{Discussion and Conclusion}

\subsection{Discussion}
Seed-based topological filtering achieves over 80\% state space reduction on furniture-rich indoor scenes without semantic labels or scene-specific training. By construction, every state in $\mathcal{S}$ is reachable from the seed, guaranteeing that all path queries within $\mathcal{S}$ are solvable---the 100\% success rate in Tables~\ref{tab:planning} and~\ref{tab:pct_plaza50} is therefore a structural property of the extracted surface, not an empirical coincidence. This guarantee comes at the cost of completeness: regions separated from the seed by physical barriers (e.g., closed doors) are excluded, producing false negatives that multi-seed initialization could mitigate. The degree of state space reduction is environment-dependent, ranging from 43.8--67.4\% on the open-structure PCT scenes to 81.3--89.5\% on the furniture-rich Matterport3D scenes, as the filtering benefit scales with the volume of geometrically valid but physically unreachable structures.

\textbf{Parameters.} All four parameters ($r$, $T_{\text{conn}}$, $H_{\text{clear}}$, $R_{\text{inf}}$) transferred without per-scene tuning across all eight scenes; $T_{\text{conn}}$ should exceed the tallest expected stair riser ($\approx$0.2--0.3\,m).

\textbf{Limitations.} The framework assumes a static occupancy map; incremental update mechanisms for dynamic environments remain future work. Ground support filtering assumes dense surface reconstruction and may yield incomplete surfaces on sparse outdoor LiDAR data. The method also requires correct seed initialization; although nearest-valid-candidate selection proved robust across all tested scenes, pose discretization errors could in principle place the seed in an unintended connected component.

\subsection{Conclusion}
Topological reachability proves an effective proxy for physical traversability, resolving geometric ambiguities across the eight indoor and semi-outdoor scenes evaluated. The resulting state space reduction (80\%+ on furniture-rich indoor scenes) enables sub-millisecond A* search on the Matterport3D scenes without per-scene parameter tuning. Future work targets dynamic environments via incremental map updates, as well as validation on real-world robotic platforms.

\section*{acknowledgement}
This work was financially supported by the National Key R\&D Program of China (Grant No. 2024YFA1211400) and Shenzhen Science and Technology Program (Grant No. ZDCY20250901112804006).

\bibliographystyle{IEEEtran}
\bibliography{references}

\end{document}